\documentclass[letterpaper, 10 pt, conference]{ieeeconf}  

\IEEEoverridecommandlockouts                              

\usepackage{amsmath} 
\usepackage{amssymb}  
\usepackage{epsfig} 
\usepackage{fancyhdr}
\usepackage{graphicx,xcolor}
\usepackage{lipsum}
\usepackage[font={footnotesize}]{caption}
\usepackage[caption=false]{subfig}

\def\bfq{{\mbox{\boldmath $q$}}}

\def\bfx{{\mbox{\boldmath $x$}}}

\def\bfR{{\mbox{\boldmath $R$}}}

\usepackage{moreverb,url,algorithm}

\usepackage{subfig}

\def\Scenario{\centerline{\includegraphics[scale=0.4]{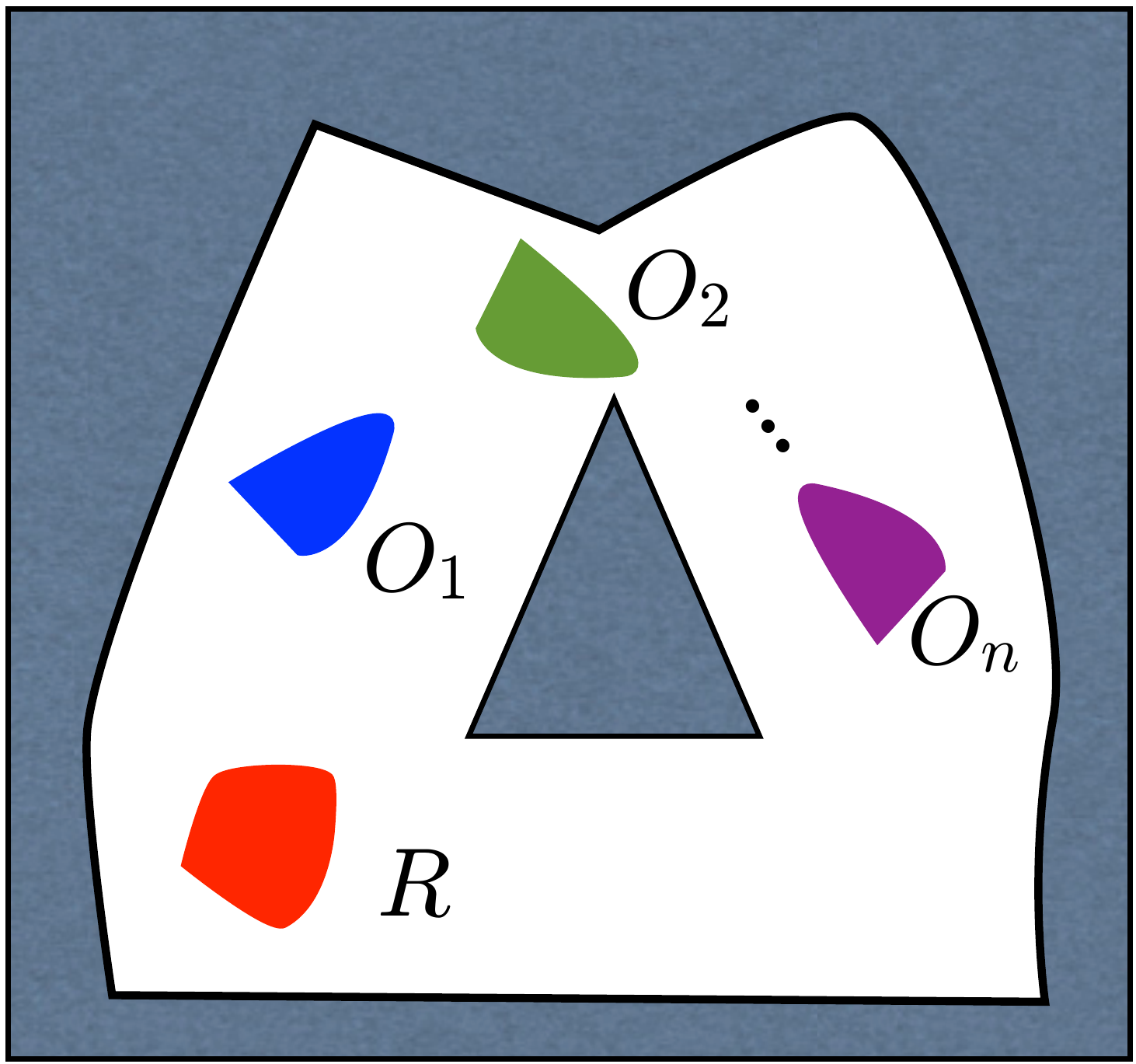}}}
\def\CStruct{\centerline{\includegraphics[scale=0.45]{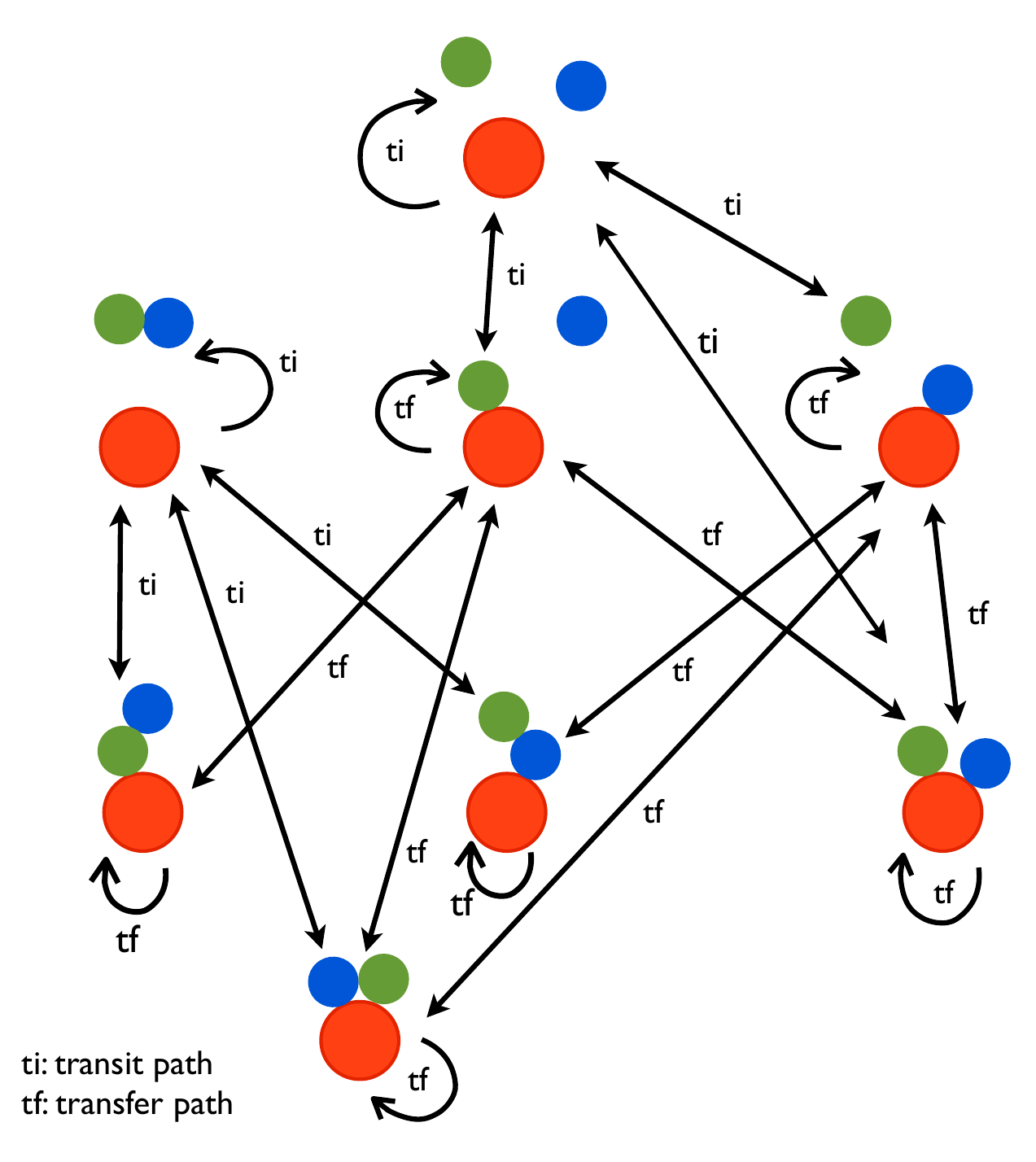}}}
\def\CStructOne{\centerline{\includegraphics[scale=0.3]{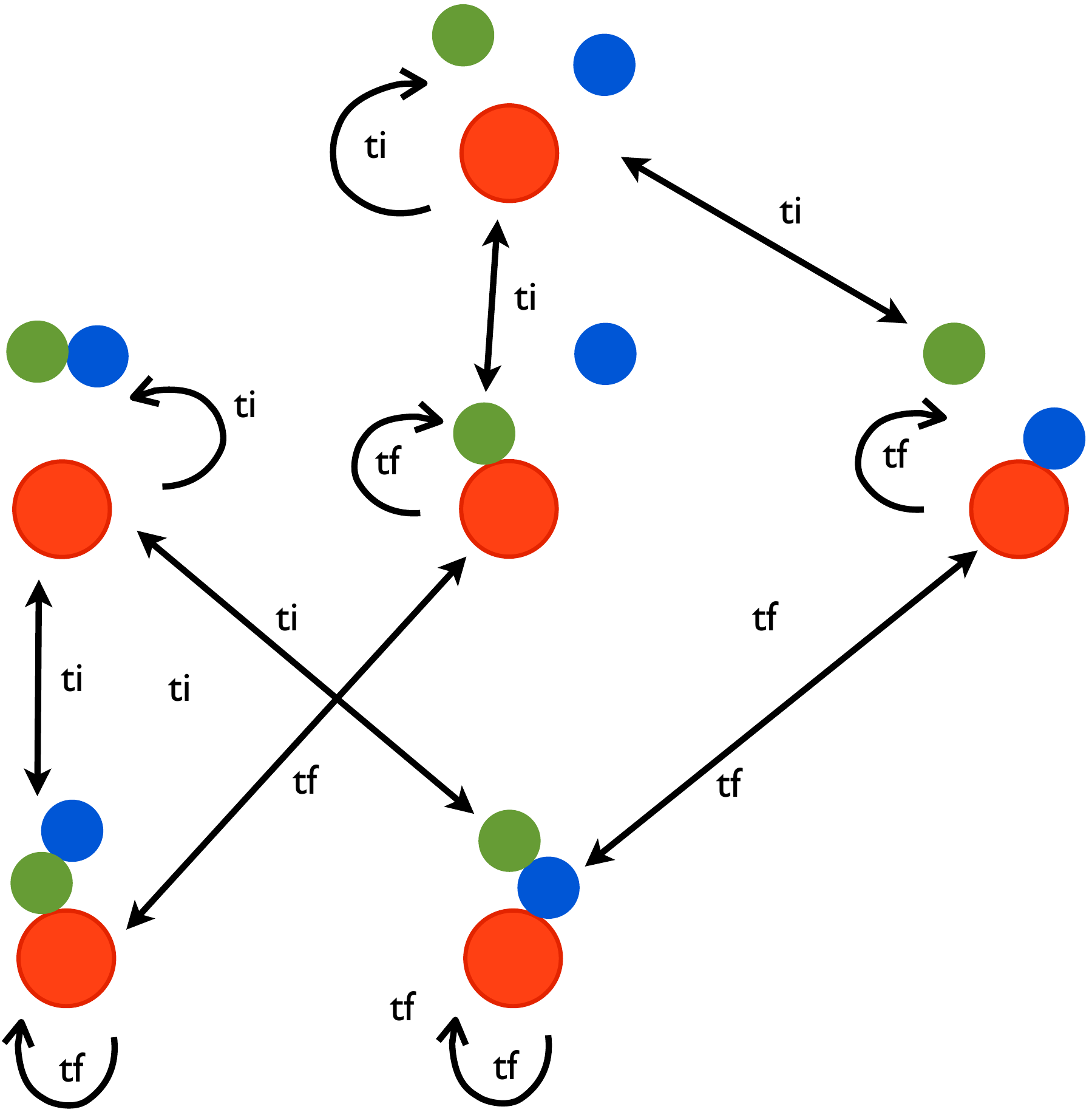}}}
\def\StratConf{\centerline{\includegraphics[scale=0.6]{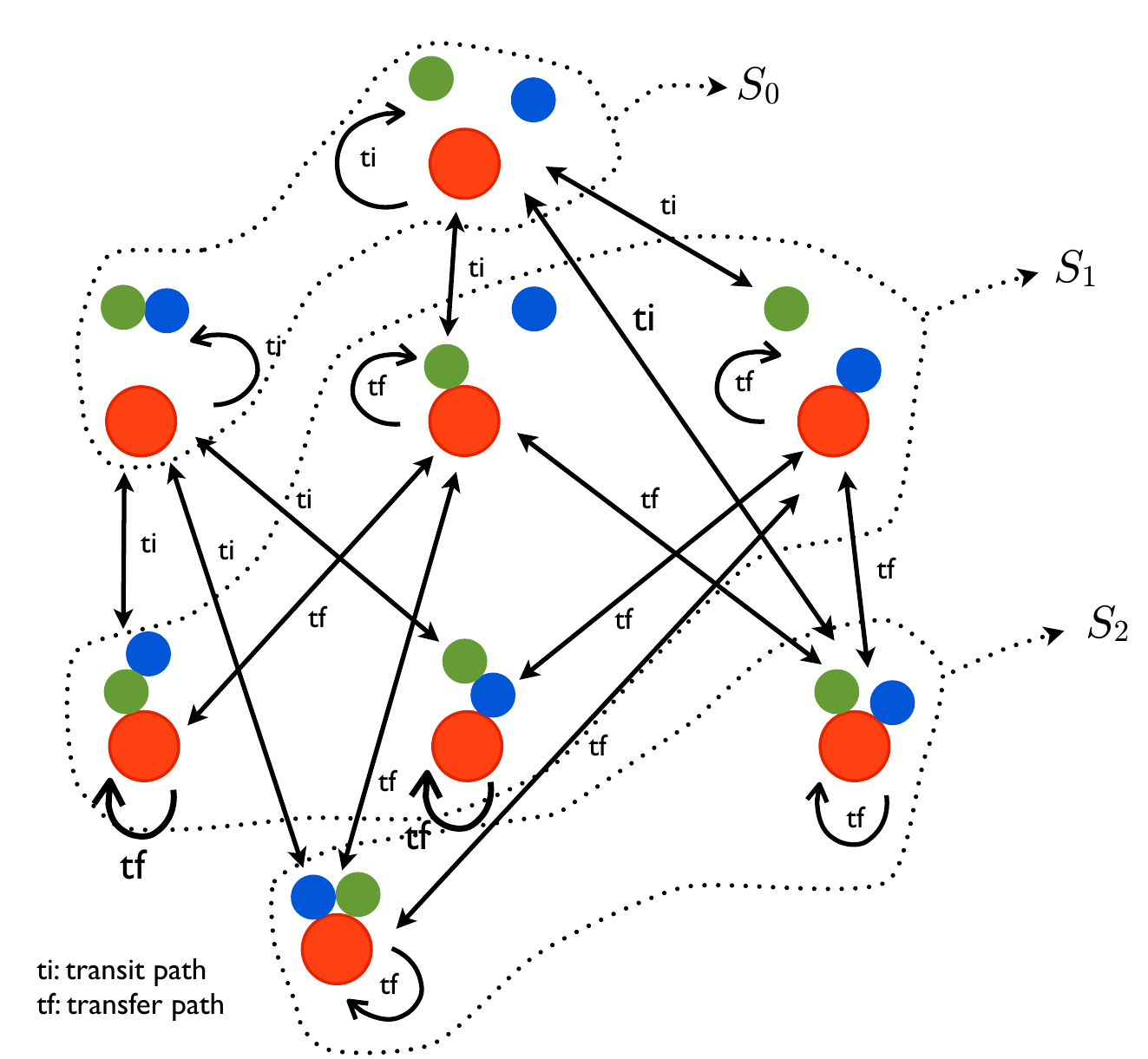}}}
\def\Ex{\centerline{\includegraphics[scale=0.3]{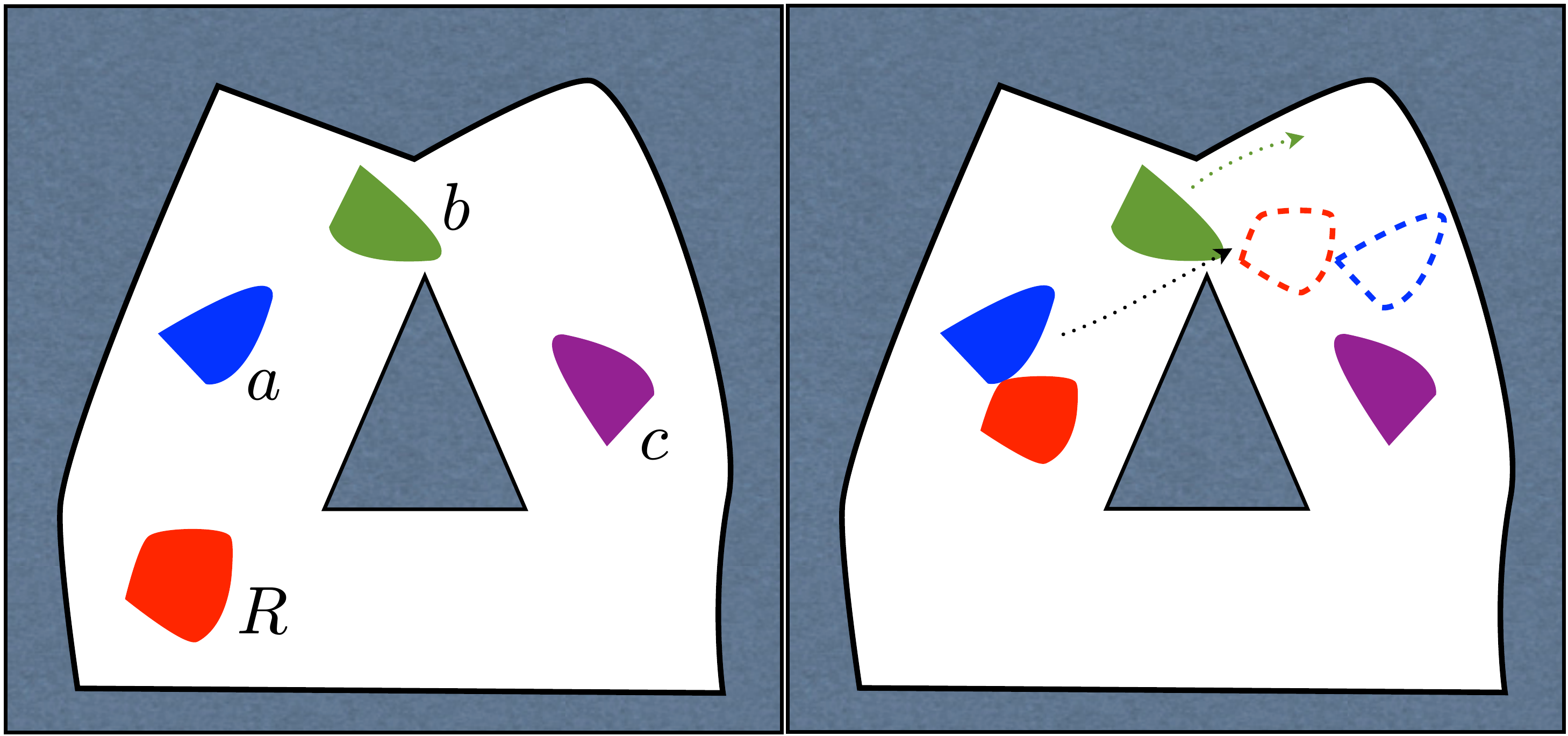}}}
\def\ExOne{\centerline{\includegraphics[scale=0.3]{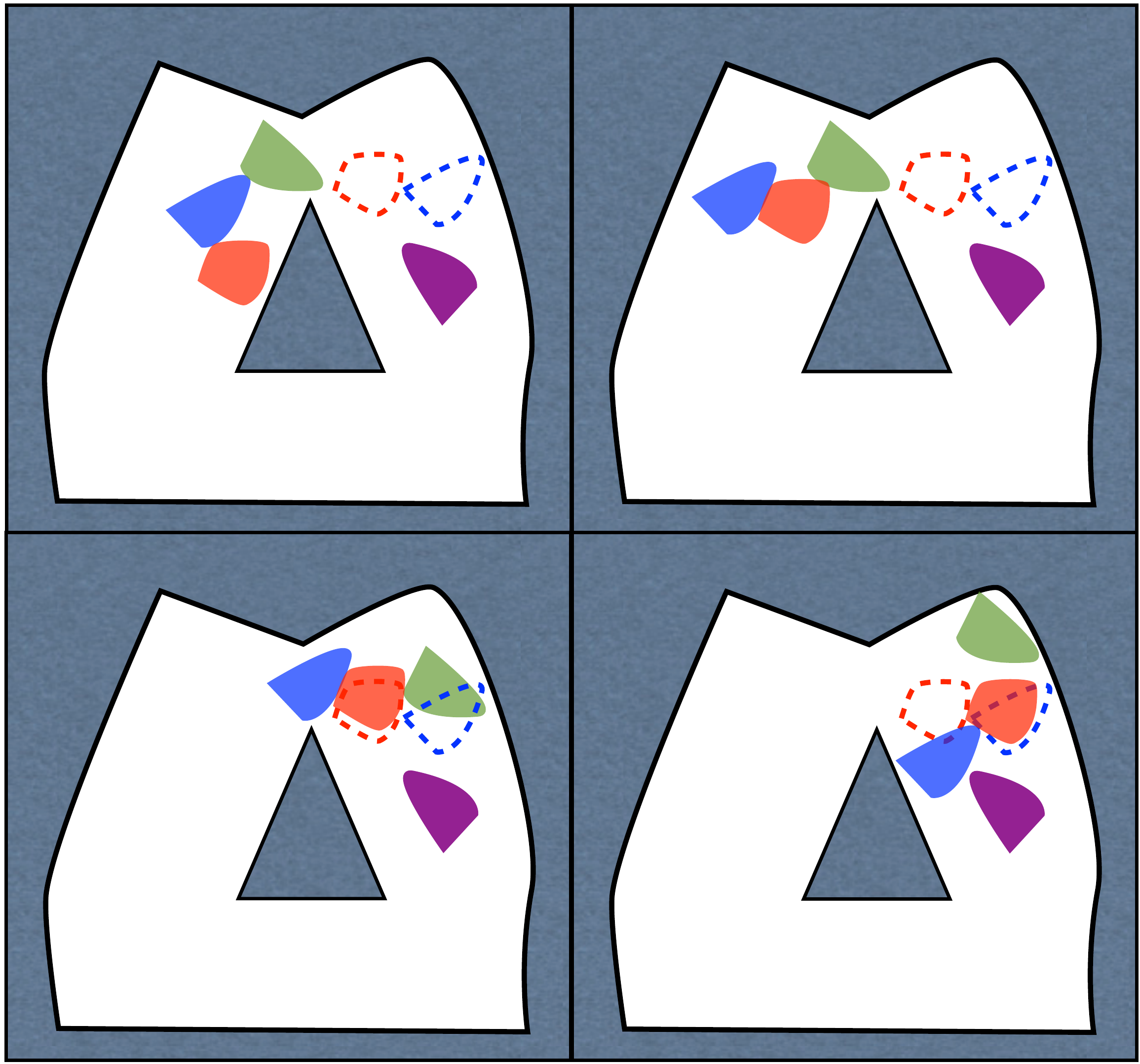}}}

\title{\LARGE \bf
	Decidability in Robot Manipulation Planning
}

\author{Marilena~Vendittelli$^{1}$, Jean-Paul~Laumond$^{2}$, Bud~Mishra$^{3}$
	\thanks{$^{1}$Dipartimento di Ingegneria dell'Informazione, Elettronica e Telecomunicazioni, via Eudossiana 18, 00184 Roma, Italy. E-mail: marilena.vendittelli@uniroma1.it}
	\thanks{$^{2}$CNRS, LAAS, Toulouse, France}
	\thanks{$^{3}$Courant Institute, NYU, New York, US}
}

\begin{document}
\maketitle

\begin{abstract}
Consider the problem of planning collision-free motion of $n$ objects in the plane movable through contact with a robot that can autonomously translate in the plane and that can move a maximum of $m \leq n$ objects simultaneously. This represents the abstract formulation of a manipulation planning problem that is proven to be decidable in this paper. The tools used for proving decidability of this simplified manipulation planning problem are, in fact, general enough to handle the decidability problem for the wider class of systems characterized by a stratified configuration space. These include, for example, problems of legged and multi-contact locomotion, bi-manual manipulation. In addition, the described approach does not restrict the dynamics of the manipulation system to be considered. 
\end{abstract}


\section{Introduction}
The problem of planning collision free motion for a free-flying single-body robot in environments populated by static obstacles has been widely studied in the past decades and can be considered today well understood.
In this paper we consider a generalization of this basic problem by allowing the presence of {\em movable obstacles}, i.e., objects in the environment that the robot can move by ``grasping'' them, while avoiding collisions with all the obstacles and objects.

The problem of motion planning in the presence of movable obstacles was first introduced in~\cite{Wi:88}, the corresponding journal version appearing in~\cite{Wi:91}, where the decidability is proven for the case of discrete grasps.
This problem was further generalized in~\cite{AlSiLa:89} to the so-called manipulation planning problem where the movable obstacles are considered as objects to be moved to reach a goal position. In that paper the authors present an algorithm for the case of discrete placements and grasps. This is the formulation briefly described in Chapter 11 of Latombe's book~\cite{La:91}.
Decidability of the problem in the case of continuous grasps and placements was shown in~\cite{DaLaAl:92} considering one movable object.

While~\cite{BeStKuLiMa:10} provides an efficient probabilistically complete algorithm in the case of several movable obstacles, the decidability problem, i.e., the existence of an exact algorithm that decides wether a solution exists in finite time, remained open even in the case of two movable objects as also mentioned in~\cite{KaBoKo:04}.

In this paper we first prove that the manipulation planning problem is decidable for a robot that can freely translate in the plane and manipulate $n$ objects that can move only if they are in contact with the robot. The proof is based on a cell decomposition of the collision-free contact configuration space and on a {\em reduction property}. This property establishes the equivalence of two types of paths: namely, paths continuously satisfying the contact constraint and manipulation paths, along which the objects either translate rigidly with the robot as a single object ({\em transfer paths}) or remain in a fixed position while the robot moves freely ({\em transit path}). To prove that the reduction property holds for the considered manipulation model we make use of the controllability result in~\cite{GoBu:01}.

The decidability procedure for this simplified case of manipulation planning problem uses methodological tools which are, in fact, abstract enough to allow handling a more general class of manipulation and planning problems.

The result, presented here, lays the basis for answering important questions such as under which conditions motion in contact can be reduced to a manipulation path, how to efficiently construct manipulation graphs related to many different problems (climbing, walking, multi-contact planning), and how to determine the rate of convergence of probabilistic planners for the manipulation of multiple objects.

This paper is an extension of our previous work (first presented at~\cite{VeLaMi:14}, then published~\cite{VeLaMi:15}) in which initially we had only proved the decidability for 3 disks. The approach was based on a three-stage paradigm: decomposition of the collision-free composite configuration space of robot and objects, retraction of the decomposition on the boundary of the collision-free grasping space, construction of the manipulation graph using the small space stratified controllability concept to define nodes and arcs of the graph. 
A very similar approach has then been applied in~\cite{DePaLo:16} to prove the decidability of the prehensile task and motion planning classes of problems.

As a matter of facts, the approach introduced in~\cite{VeLaMi:14} was general enough to include a wide class of planning problems. In the present paper we show how the same approach can be further generalized to consider an arbitrary number of movable objects with arbitrary shape and discuss the general characteristics of the approach which permits dealing with other classes of planning problems.

The paper is organized as follows. The next section formalizes the problem after defining the configuration space and its connectivity through manipulation paths. Section~\ref{Sect:Controllability} establishes the conditions under which motion in contact can be reduced to a manipulation path. Section~\ref{Sect:ManGraph} illustrates the main steps for the construction of the manipulation graph. Section~\ref{sect:genapp} opens perspective about the generalization of the results to other manipulation planning problems and finally, Sect.~\ref{Sect:Conclusion} concludes the paper. 

\section{Problem formulation}
\label{Sect:PbFor}

\begin{figure}[h!]
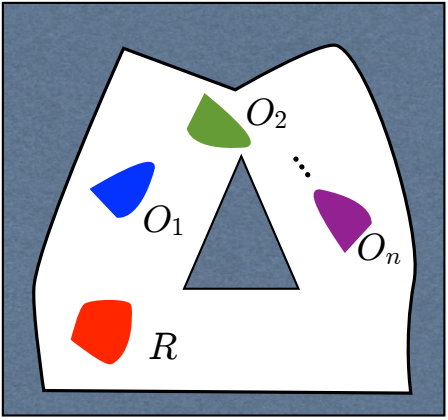

\begin{center}
\Scenario
\caption{Scenario of the considered manipulation planning problem discussed here.}
\label{Fig:Scen}
\end{center}
\end{figure}

\noindent
Consider the scene in Fig.~\ref{Fig:Scen}: the ``robot'' $R$ can translate autonomously in a polygonal (or semi-algebraic) environment populated by fixed obstacles and objects $n$ objects $O_1, \, O_2, \, \dots, \, O_n $ that $R$ can move by establishing a contact with them. More specifically, the objects $O_1, \, O_2, \, \dots, \, O_n $ translate rigidly with the robot when in contact with it; otherwise, they are considered as fixed obstacles.
\subsection{Configuration space}
\noindent
The configuration spaces of the robot and the objects are defined as: 
\begin{itemize}
\item[{\tiny $\bullet$}] ${\cal C}_R = \bfR^2$, the configuration space of the robot; 
\item[{\tiny $\bullet$}] ${\cal C}_{O_i} = \bfR^2$ the configuration space of $O_i$, $i \in {1, \dots, n}$.
\end{itemize}
The combined configuration space is obtained as ${\cal C}={\cal C}_R \times {\cal C}_{O_1} \times {\cal C}_{O_2} \times \dots \times {\cal C}_{O_n} = \bfR^{2n}$.
A configuration $\bfq \in {\cal C}$ is given by the $n$-tuple  $\bfq = (\bfq_R, \bfq_{O_1}, \bfq_{O_2}, \dots, \bfq_{O_n})$, where $\bfq_R \in {\cal C}_R$, $\bfq_{O_i} \in {\cal C}_{O_i} $, $i \in {1, \dots, n}$.

The collision-free configuration space ${\cal C}_{\mbox{free}}$ is obtained by removing from $\cal C$ the set of inadmissible configurations:
\begin{itemize}
\item[{\tiny $\bullet$}] $\bfq_R$ such that the robot is in contact with static obstacles or overlaps with either static or movable obstacles;
\item[{\tiny $\bullet$}] $\bfq_{O_i}$, $i \in {1, \dots, n}$ such that $O_i$ overlaps with the static obstacles, the robot or with ${O_j}$, $j\neq i$. Note that contact between objects and obstacles is allowed.
\end{itemize}

\subsection{Configuration space paths and manipulation paths}
\label{Sect:PlanStruct}

\noindent
Configuration space paths may or may not include contacts. To move the objects, however, the robot must be in contact with the objects. Paths of interest in ${\cal C}$ can be categorized according to the two motion modalities:
\begin{itemize}
\item[{\tiny $\bullet$}]  robot free motion: this is a path in ${\cal C}$ characterized by the absence of contact between the robot and the objects;
\item[{\tiny $\bullet$}]  contact motion: this is a path in ${\cal C}$ constrained by the condition that the robot is in contact with at least one object; along the path the robot position and the positions of the objects relative to the robot can change.
\end{itemize}

The above described paths might or might not be feasible for a manipulation system depending on its characteristics.
In the following developments we consider only manipulation by rigid grasp. 
Therefore, not all the configuration space paths are feasible in our setting. Feasible motions correspond to paths of two types: 
\begin{itemize}
\item {\em transfer paths} along which the robot grasps at least one object and moves rigidly with it (the objects not grasped remain in a fixed position); along these paths the relative configurations between robot and objects in contact do not change.
\item {\em transit paths} along which the robot moves alone and the objects remain in fixed positions. 
\end{itemize}
A sequence of transit and transfer paths is called a {\em manipulation path.} 

\begin{figure}[t]
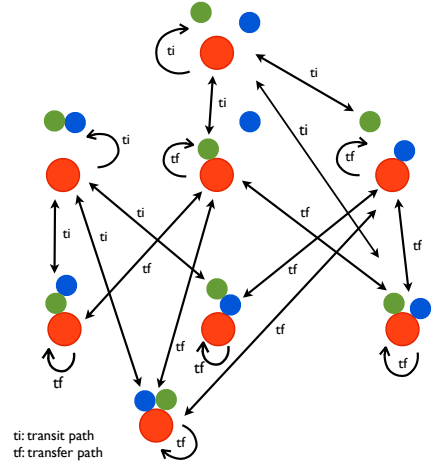
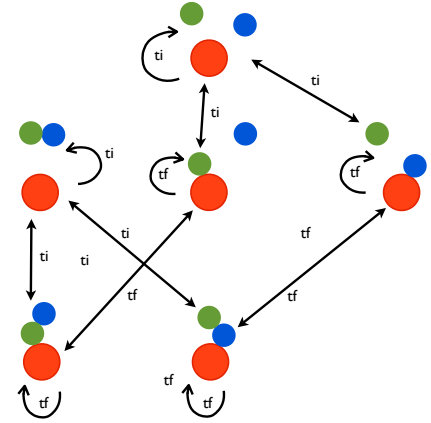

\subfloat[][]{\CStruct \label{fig:CStruct}}  
\qquad \quad
\subfloat[][]{\CStructOne \label{fig:CStructOne}}
\caption{Structure of ${\cal C}$ induced by the contact constraints and interconnection of the contact submanifolds through transit and transfer paths. The case illustrated in~\protect\subref{fig:CStruct} corresponds to $n=2$ and $m=2$, while in~\protect\subref{fig:CStructOne} it is considered $n=2$ and $m=1$.}
\label{fig:CStructOneTwo}
\end{figure}

\subsection{Configuration space connectivity through manipulation paths}
\label{Sect:ContactSubM}
\noindent
The number $m$ of objects that the robot can manipulate at the same time is in general limited and it affects the structure of ${\cal C}$ induced by the contact constraints. 
In this section we illustrate the structure of ${\cal C}$ in terms of the submanifolds defined by the contact constraints and their interconnection through transit and transfer paths. 

Figure~\ref{fig:CStructOneTwo} shows representative configurations in each submanifold for the cases $n=2$ and $m=2$ or $m=1$, respectively illustrated in Fig.~\ref{fig:CStruct} and Fig~\ref{fig:CStructOne}. 
The embedding configuration space ${\cal C}$ has dimension 6 and foliates with the position of the movable objects. In particular, leaves of dimension 2  correspond to fixed positions of the two objects. Transit paths belong to one of these leaves. A representative configuration in this manifold is shown at the top of Fig.~\ref{fig:CStruct}. Manipulation paths across the leaves may require leaving the manifold. 

Configurations on the second row (from top) of Fig.~\ref{fig:CStruct} represent the single-contact manifold which has dimension~5. The leaves of interest for the problem of interest have dimension 3 and correspond to fixed positions of the object which is not in contact with the robot. Manipulation paths across the leaves (change of the position of the object not in contact) require leaving the submanifold.

The double-contact submanifold, represented by the configurations on the third row of Fig.~\ref{fig:CStruct}, has dimension 4 and foliates with the relative position of the contact points. The leaves of interest have dimension 3 and 2 and correspond respectively to one or both the points of contact being fixed. Manipulation paths across the leaves (change of the contact point) require leaving the submanifold.

Finally, the triple-contact submanifold has dimension 3, it foliates with the position of the contact points and the leaves have dimension 2. Manipulation paths across the leaves require leaving the submanifold.

This last submanifold is not present in the structure of ${\cal C}$ when $m=1$ as illustrated in Fig.~\ref{fig:CStructOne}. However, our solution requires considering also this submanifold of ${\cal C}$ to prove decidability, as will be illustrated in what follows.

In general, the structure of $\cal C$ depends both on the number $n$ of movable objects and on $m$, the maximum number of objects that the robot can move at the same time.

As will be illustrated in section~\ref{sect:StratContr}, the ``manipulability'' properties associated with the above described submanifolds are actually transversal to this geometric structure and depend on the controllability of the underlying manipulation system (see Sect.~\ref{Sect:ContManipSys}). 

\subsection{The manipulation planning problem}
Given the definitions and analysis in the previous sections, we can formulate the following problem.

\noindent
{\em Manipulation Planning Problem.} Assigned an initial configuration $\bfq_s \in {\cal C}_{\mbox{free}}$ and a goal configuration $\bfq_g \in {\cal C}_{\mbox{free}}$, find a sequence of transit and transfer paths joining $\bfq_s$ to $\bfq_g$, if it exists.

To prove that this problem is decidable we adopt the same approach as~\cite{DaLaAl:92}. First we study the problem of reducing  configuration space paths belonging to the contact submanifolds represented in Fig.~\ref{fig:CStructOneTwo} to manipulation paths. Then, we determine a cell decomposition of the contact space. Finally, we construct the manipulation graph whose connected components characterize the existence of solutions to the manipulation problem defined above. In case $m < n$ we remove some nodes and the corresponding arcs from the manipulation graph. 

The first part of our approach consists, in fact, in answering the following question: is it possible to reduce any collision-free configuration space path describing motion of the robot in contact with at least one object to a (finite) sequence of transit and transfer paths?
Answering this question requires studying the local controllability of the dynamic system that is possible to associate with the manipulation model. 
The analysis is described in the following section and makes use of the result by Goodwine and Burdick~\cite{GoBu:01} providing sufficient conditions for controllability of kinematic control systems on stratified configuration spaces. 

\section{Controllability of the manipulation system}
\label{Sect:Controllability}
To answer the first part of the manipulation planning problem, we define here the simple kinematics describing the manipulation system underlying the planning problem under consideration.
This system has a {\em stratified configuration space} and we use the result in~\cite{GoBu:01}  to establish its  local controllability. 

\subsection{Controllability definitions}
This section recalls the controllability definitions of interest to prove decidability of the manipulation problems of interest.
Given an open set $V \subseteq M$, let $R^V(x_0, T)$ be the set of states $x_f$ such that there exists $u: [0, T] \rightarrow {\cal U}$ that steers the control system from $x(0) = x_0$ to $x(T) = x_f$ and satisfies $x(t) \in V$ for $0 \leq t \leq T$, where $\cal U$ is the set of admissible control inputs. Define the set of states reachable up to time T as
\begin{equation}
\label{eq:ReachSet}
R^V(x_0, \leq T) = \bigcup_{0 < \tau \leq T} R^V(x_0, T).
\end{equation}

\noindent
{\em Small Time Local Controllability (STLC)}: A smooth analytic system is {\em small time locally controllable (or STLC)} if $R^V(x_0, \leq \tau)$ contains a neighborhood of $x_0$ for all neighborhoods $V$ of $x_0$ and $T > 0$.

A more ``natural'' notion of controllability for proving decidability is the so-called local--local controllability (LLC) notion introduced in~\cite{HaHe:70}.

\noindent
{\em Local--Local Controllability (LLC)}: A smooth analytic system is {\em local--local controllable (or LLC)} from $x_0$ if $\forall \epsilon > 0 \,\,\exists \,\delta(\epsilon) > 0$ such that for all admissible states $x, \,\,\, || x - x_0|| < \delta(\epsilon)$ there exists admissible control $u: [0, T] \rightarrow {\cal U}$, producing the state trajectory $x(t), \, t \in [0,T]$ with
\begin{equation}
x(t_1)=x_0, \,\, x(t_2)=x \,\, \mbox{\rm  and } \,\, ||x(t)-x_0||<\epsilon, \,\, \forall t \in [t_1, t_2].
\end{equation}

LLC requires that the trajectory to reach points in the neighborhood of a given state $x_0$ is local. The time, however, is not specified (or bounded) in advance.
For systems with bounded inputs STLC and LLC are equivalent properties.

\subsection{Controllability on stratified configuration spaces}
\label{Sect:ContManipSys}

The definition of STLC above is generalized in~\cite{GoBu:01} to include the case of stratified systems.
We briefly recall here the main definitions and properties of stratified configuration spaces and the stratified controllability property that we prove to hold in our case. 

\noindent
{\em Stratified configuration manifold} (Definition 2.2 in~\cite{GoBu:01}): Let $M$ be a manifold (possibly with boundary), and $n$ functions $\Phi_i$: $M  \mapsto \mathbb{R}$, $i=1,\dots,n$ be such that the level sets $S_i = \Phi_i^{-1}(0) \subset M$ are regular submanifolds of $M$, for each $i$, and the intersection of any number of the level sets, $S_{i_1i_2\dots i_m} =  \Phi_{i_1}^{-1}(0) \cap  \Phi_{i_2}^{-1}(0) \cap \dots  \Phi_{i_m}^{-1}(0)$, $m \leq n$ , is also a regular submanifold of $M$. Then $M$ and the functions $\Phi_{i}$, define a {\em stratified configuration space}.

The driftless systems defined on stratified configuration manifolds are described on each stratum, or on strata  intersections, by equations of motion characterized by smooth vector fields and the only discontinuities present in the equations of motion are due to transitions on and off the strata or their intersections.

\noindent
{\em Stratified controllability} (Definition 3.2 in~\cite{GoBu:01}): A stratified system is {\em stratified controllable} in $S_I$ from $x_0 \in S_I$ if $R^V(x_0, \leq T)$ contains a neighborhood of $x_0$ in $S_I$ for all neighborhoods $V \subseteq S_I$ of $x_0$ and $T > 0$, where $R^V(x_0, \leq T)$ is defined by Eq.~(\ref{eq:ReachSet}) with $V \subseteq S_I$.

\noindent
{\em Stratified controllability} (Proposition 4.4 in~\cite{GoBu:01}): if there exists a nested sequence of submanifolds at the configuration $x_0$
$$x_0 \in S_p \subset S_{p-1} \subset \dots \subset S_1 \subset S_0 = M,$$
where the subscript is the codimension of the submanifold,
such that the associated involutive distributions satisfy
$$\sum_{j=0}^p \stackrel{\relbar}{\Delta}_{S_j}|_{x_0} = T_{x_0}M$$
and each $\stackrel{\relbar}{\Delta}_{S_j}$ has constant rank for some neighborhood $V_j \subset S_j$,
of $x_0$, then the system is stratified controllable from $x_0$ in $M$.

Stated differently, if the involutive closures of the distributions associated to each submanifold in the nested sequence intersect {\em transversely} then the system can flow in any direction in $M$. 
Intuitively, this controllability concept allows to prove that any path in contact in the composite configuration space of the robot and the movable objects can be approximated by a sequence of transit and transfer paths, i.e., a path in the stratified configuration space of the manipulation system.

\subsection{Stratified controllability of the manipulation system}
\label{sect:StratContr}

\noindent
To use the stratified controllability concept for proving that a contact path can be transformed into a manipulation path, we consider ambient manifolds and stratifications defined by the contact submanifolds.
In particular, there exist $n_p= \left(
\begin{array}{c}
  n      \\
  p      
\end{array}
\right)$, $p = \{1, \dots, m\}$, ambient submanifolds $M_{n_p}$ given by the combined configuration space of the robot and the $p$ objects in contact. The dimension of each ambient submanifold is equal to ${\rm dim}(M_{n_p})= 2 + 2 p$ and it contains $\left(
\begin{array}{c}
  p      \\
  p-i      
\end{array}
\right)$
submanifolds of codimension $p-i$, $i = \{0, \dots, p-1\}$ defined by all the possible contact combinations of the $p$ objects with the robot.
Note that these submanifolds are leaves of the combined configuration space $\cal C$ of the robot and all the $n$ obstacles.

The stratified controllability property  is easily verified to hold on each of these ambient manifolds, while it is not verified on submanifolds of $\cal C$ of dimension greater than $2 + 2 p$. Hence, the considered manipulation system is stratified controllable on all the leaves of $\cal C$ defined by the submanifolds $M_{n_p}$.
How to build the nested sequence of submanifolds providing stratified controllability is the subject of what follows. 

Denote by $\bfx = (x_R, y_R, x_{O_{c_1}}, y_{O_{c_1}}, \dots, x_{O_{c_p}}, y_{O_{c_p}})^T$, a configuration of the manipulation system formed by the robot and $p$ movable objects. Set the maximum number of objects that the robot can move at the same time equal to $p$. The equations of motion on each stratum are determined by considering that $R$ can only translate in the plane and the objects can be moved when in contact with $R$ with a stable grasp. The control system underlying this manipulation model can be written in the form 
\begin{equation}
\dot \bfx = g_1^{S_i} u_1 + g_2^{S_i} u_2,
\label{eq:ManipContrSys}
\end{equation}
where $u_1$, $u_2$ are the robot cartesian velocities considered as the system inputs and $g_1^{S_i}$, $g_2^{S_i}$ are the input vector fields that have a different expression on each substratum.
In particular, in $S_0 ={\cal C}$ we have
\[
g_1^{{\cal S}_0} = (1,\,0,\,{\bf0})^T, \quad  g_2^{{\cal S}_0} = (0,\,1,\,{\bf0})^T,
\]
where the first two components of the vector fields determine the motion of the robot and are invariant on all the strata.
In $S_0$ they describe the free motion of the robot alone on a leaf of $\cal C$ that depends on the position of the objects. The remaining components have been grouped in vectors of zeros of dimension $p-2$.

On the single-contact manifolds ${\cal C}_{1}$ the input vector fields have the same first two components, the $(i+2)$-th entry of $g_1^{{\cal C}_{1}}$ and the $(i+3)$-th entry of $g_2^{{\cal C}_{1}}$, $i \in {1,\dots,p}$, equal to 1, the remaining components equal to zero.
Flowing along these vector fields amounts to moving the object in contact while staying on a leaf that depends on the position of the objects that are not touched by the robot.
Since all the single-contact manifolds have codimension 1, $S_1$ will be equal to either one of them in the sequence of nested submanifolds. This control will however implicitly assume that the position of the other objects will remain constant, i.e., the system is flowing on a leaf of the single contact manifold.
Hence, if the system is stratified controllable, it will be only possible to prove that any path in contact on a leaf of $S_1$ can be approximated by a manipulation path. In fact, we need this local controllability to guarantee that a collision free path in continuous contact can be approximated by a manipulation path that is contained in a neighborhood of the original contact path. 
Complete decidability study requires the analysis of the manipulation graph connectivity.

The vector fields describing the motion on the subsequent strata can be defined iteratively by setting the components corresponding to the coordinates of the objects in contact equal to 1 and all the remaining components equal to zero. As noted above, the first two components of each vector field do not change across the strata and only one stratum for each codimension will be considered in the sequence of nested submanifolds. Analogously to the single contact case, this implies that the reduction property, or the possibility to approximate a collision-free path with a manipulation path, is only valid on a leaf of each substratum $S_j$, $j \in {1, \dots, m}$. Note that, in our setting, a substratum $S_j$ collects all the contact submanifolds of codimension $j$ defined by the contact of the robot and $j$ of the $n$ movable obstacles.

As an example, assuming that $n=m=2$, considering the unique ambient submanifold $M_{n_p}$ with $n=p=2$, on the single-contact manifolds ${\cal C}_{1}$, composing the stratum $S_1$,  the input vector fields in~(\ref{eq:ManipContrSys}) have the expressions
\[
g_1^{S_1} = (1,\,0,\,1,\,0,\,0,\,0)^T, \quad  g_2^{S_1} = (0,\,1,\,0,\,1,\,0,\,0)^T
\]
or
\[
g_1^{S_1}  = (1,\,0,\,0,\,0,\,1,\,0)^T, \quad  g_2^{S_1}  = (0,\,1,\,0,\,0,\,0,\,1)^T.
\]

On the double-contact manifold $S_2 = {\cal C}_{2}$, in case of contact with the objects $O_1$ and $O_2$, it is
 \[
g_1^{S_2} = (1,\,0,\,1,\,0,\,1,\,0)^T, \quad  g_2^{S_2}  = (0,\,1,\,0,\,1,\,0,\,1)^T.
\]
On this stratum the objects move with the robot without changing the points of contact.

It is easy to verify that the stratified controllability proposition holds by choosing as involutive distributions
$$\stackrel{\relbar}{\Delta}S_2=\mbox{span }(g_1^{S_2} \,\,\, g_2^{S_2})$$
$$\stackrel{\relbar}{\Delta}S_1=\mbox{span }(g_1^{S_1}  \,\,\, g_2^{S_1})$$ 
$$\stackrel{\relbar}{\Delta}S_0=\mbox{span }(g_1^{S_0}  \,\,\, g_2^{S_0})$$
where $g_1^{S_1}$ and $g_2^{S_1}$ can have either one of the expressions  provided above.

\begin{figure}[htbp]
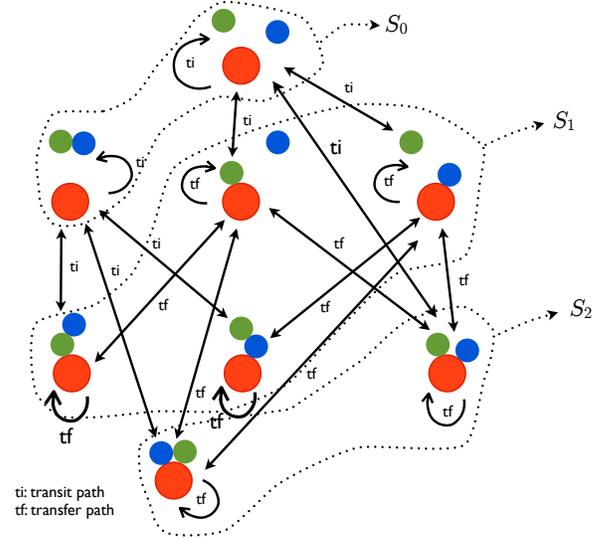

\begin{center}
\StratConf
\caption{Stratification of the configuration space induced by the robot-objects contact constraints in the case $n=m=2$. Note how the relevant contact submanifolds needed to prove controllability are determined by the contact configurations between the robot and the objects.}
\label{fig:StratConf}
\end{center}
\end{figure}

Figure~\ref{fig:StratConf} illustrates the stratification of the configuration space induced by the contact constraints in the case $n=m=2$. 
By virtue of the controllability property described above, any continuous path in contact with the robot in $S_2$ can be approximated by a manipulation path. 

Considering then as ambient manifold one of the two submanifolds $M_{n_p}$ with $n=2$ and $p=1$ defined by the object in contact with the robot. The preceding construction can be adapted to prove stratified controllability on each of these submanifolds that are leaves of the complete configuration space defined by the robot and the two movable objects. Hence, a path in contact in each leaf of $S_1$ can be approximated by a manipulation path that possibly goes through strata of lower codimension. 

Until here we are not considering obstacles, hence the existence of a manipulation path depends locally on the controllability property. We take the hypothesis that the free configuration space of the robot without considering the objects is an open (same as the {\em hypothesis H} in~\cite{DaLaAl:92}). Then, any collision-free path in contact can be approximated by a collision-free manipulation path. Intuitively, if the system is controllable, the ``maneuvers'' involved in the manipulation path can be made as small as desired so as the manipulation path remains in the neighbourhood of the contact path.

It is very easy to prove that in the general case of $n$ movable objects and $m$ maximum number of objects that the robot can manipulate at the same time, the reduction property is valid in each leaf of the manifolds of codimentions from $m$ to 1. This property will be used to prove decidability by showing that if a solution exists it will either lie on a leaf of the free configuration space defined by a fixed position of the objects, i.e., the solution does not imply manipulation of the objects, or it passes through the submanifolds defined by the contact constraints between the robot and the obstacles. On each leaf of these submanifolds the reduction property holds, then a manipulation path exists if start and goal configurations can be connected through collision-free transit paths and paths in contact. 

\section{Building the manipulation graph}
\label{Sect:ManGraph}
The reduction property established in the previous section leads to the conclusion that any collision-free path in contact contained in any leaf of $\cal C$ defined by the ambient manifolds $M_{n_p}$ and in any leaf of the submanifolds defining the nested sequence supporting the controllability property is equivalent to a manipulation path.
The remaining key issue  involves building a geometric data structure that accounts for the obstacles presence and ultimately for the decidability of the manipulation problem. 

We propose here an extension of the manipulation graph as it has been introduced in~\cite{DaLaAl:92} for the case of a single movable object, i.e., $n=m=1$. In that case the admissible (i.e., not in collision with static obstacles nor overlapping the object to move) contact configurations between the robot and the object were represented by the class $GRASP$.
The nodes of the manipulation graph were then given by the connected components of $GRASP$ where the controllability, and hence reduction, property is easily shown to hold. The adjacency relation was given by the existence of transit paths between two nodes.

In the case of $n$ movable objects and a maximum of $m \leq n$ objects movable at the same time by the robot, it is necessary to introduce $n_p$ classes $GRASP_{\Phi_{n_p}}$, $p=\{1,\dots,m\}$, and to build the manipulation graph over the connected components of these classes. 

Each class $GRASP_{\Phi_{n_p}}$ represents the configurations in ${\cal C}_{\rm free}$ such that the robot is in contact with the $p$ objects corresponding to one of the $n_p$ combinations also defining the ambient manifolds in Sect.~\ref{sect:StratContr} within which the stratified controllability holds. We define, hence, $\Phi_{n_p}$ as the string of length $p$ consisting in one of the $n_p$ combinations of $p$ from the $n$ objects. 
The position of the objects not included in $\Phi_{n_p}$ can change within the class. 
To distinguish the different $n_p$ combinations we introduce the notation $\Phi_{n_{p,i}}$, $i=\{1, \dots, n_p\}$. When only the length of the string $\Phi_{n_p}$ is relevant the index $i$ is not specified and we do not make a distinction between the  $n_p$ combinations.

The reduction property shown in the previous section does not apply on the whole connected components of $GRASP_{\Phi_{n_p}}$ but inside each leaf of the foliation of $GRASP_{\Phi_{n_p}}$ that keeps constant the position of the obstacles that are not in contact with the robot: any path inside these leaves can be approximated by a sequence of transit and transfer paths. These are, for example, leaves of dimension 3 in the single contact manifolds schematically represented in Fig.~\ref{fig:CStruct} for the case $n=m=2$.

The key questions are then: $(i)$ how to determine the connected components of each $GRASP_{\Phi_{n_p}}$, and $(ii)$  how to build a manipulation graph on which it is possible to decide for the existence of a manipulation path.
 
To answer the first question consider that each $GRASP_{\Phi_{n_p}}$ is by definition a contact submanifold of ${\cal C}_{\rm free}$ of dimension $2(1+n) - p$, $p=\{1,\dots,m\}$. If there exists a cell decomposition of the $2(1+n)$-dimensional space ${\cal C}_{\rm free}$, then this cell decomposition induces, by retraction on its boundary, a cell decomposition of the $(2(1+n) - p)$-dimensional contact submanifolds (up to some potential singularities we do not consider in this paper). Then, such a cell decomposition leads to a straightforward characterization of the connected components of each $GRASP_{\Phi_{n_p}}$. The first question is then reduced to the existence of an algorithm that provides a cell decomposition of ${\cal C}_{\rm free}$.  A cylindrical decomposition can be used to this aim, as proposed in~\cite{Sc:83}. 

Building the manipulation graph is the second issue to be addressed. In particular, we need a suitable adjacency relationship between the cells of the classes $GRASP_{\Phi_{n_p}}$. Note, in fact, that each cell in the $GRASP_{\Phi_{n_p}}$ includes configurations that can be joined by elementary collision-free paths. This elementary paths, however, consist in the coordinated motion of robot and objects, including those which are not in contact. In fact, 
as usual, the configuration space is unconstrained, that is, it has been constructed without considering that to change the position of an object it is necessary to move in contact with the robot. Hence, only the elementary paths that remain in the same leaf of a connected component of $GRASP_{\Phi_{n_p}}$ are guaranteed to be reducible to collision-free manipulation paths by retracting the cell decomposition of ${\cal C}_{\rm free}$ on its boundaries.

Then, we need to refine the cell decompositions of the connected components of each $GRASP_{\Phi_{n_p}}$ by considering their projections along the directions of the foliations generated by: $(i)$ transit paths (the robot moves alone), $(ii)$ transfer paths of {\em type} $n_p$ (the $ n_p$ objects are moved by the robot while the remaining $n-n_p$ do not move). 

Note first that the projection of a given cell $C_1$ onto a cell $C_2$ induces a decomposition of $C_2$ into several cells. 
The projection of a $GRASP_{\Phi_{n_{p,i}}}$ cell decomposition along the direction of its  foliations onto a $GRASP_{\Phi_{n_{p+1}}}$ gives  rise to a decomposition of this last class into multiple cells. This decomposition is further refined by projecting the cell decomposition of each $GRASP_{\Phi_{n_{p,j}}}$ such that the length of the string $\Phi_{n_{p,i}} \cup \Phi_{n_{p,j}}$ is equal to $p+1$ onto $GRASP_{\Phi_{n_{p+1}}}$. The initial cell decomposition of $GRASP_{\Phi_{n_{p,i}}}$ and $GRASP_{\Phi_{n_{p,j}}}$ can then be refined by ``lifting'' all cells in $GRASP_{\Phi_{n_{p+1}}}$ along the foliations of $GRASP_{\Phi_{n_{p,i}}}$ and $GRASP_{\Phi_{n_{p,j}}}$, respectively. 
 $GRASP_{\Phi_{n_{p+1}}}$ is then decomposed in elementary cells of which some are at the basis of two cylinders containing respectively cells of $GRASP_{\Phi_{n_{p,i}}}$ and $GRASP_{\Phi_{n_{p,j}}}$. 

The cell decomposition of $GRASP_{\Phi_{n_{p+1}}}$ may however need to be further refined.
The complete cell refinement is obtained by incrementally projecting cells of $GRASP_{\Phi_{n_{p}}}$ on cells of $GRASP_{\Phi_{n_{p+1}}}$, $p=1, \dots, m-1$, along the foliations induced by transfer paths of type $n_p$. Then each cell of $GRASP_{\Phi_{n_{m-i}}}$ is lifted to $GRASP_{\Phi_{n_{m-i-1}}}$, $i=0, \dots, m-2$.
The cells generated by this refinement procedure and belonging to at least one cylinder constitute the nodes of the manipulation graph. 

We then introduce the following adjacency relation: two cells in a class $GRASP_{\Phi_{n_{p}}}$ are adjacent by transfer paths if and only if they have a common frontier and they belong to at least one same cylinder. 

If $GRASP_{\Phi_{n_{p}}}$ belongs to a contact manifold of minimum dimension, i.e., $p=m$, and it is $n=m$ then adjacency is given by the existence of a common frontier between two cells. In fact, in this submanifold, any path in contact is equivalent to a manipulation path due to the stratified controllability property that, in this case, holds on $\cal C$. 

We can then introduce the following recursive definition of adjacency by transfer paths:\\
Two cells belonging to two different classes $GRASP_{\Phi_{n_i}}$ and $GRASP_{\Phi_{n_j}}$ are adjacent by transfer paths if and only if their projections on $GRASP_{\Phi_{n_i} \cup \Phi_{n_j}}$ along the respective foliations, induced by transfer paths, intersect cells of the class $GRASP_{\Phi_{n_i} \cup \Phi_{n_j}}$ which are adjacent by transfer paths.
Note that the recursion terminates if $m=n$ because in the $GRASP_{\Phi_{n}}$ class the adjacency is given by the existence of a common frontier between cells. 
To prove decidability the graph is always constructed by considering $m = n$. In case $m < n$, the nodes corresponding to classes $GRASP_{\Phi_{n_p}}$, $p > m$, are removed from the graph together with all adjacency relations between cells that are based on the adjacency of $GRASP_{\Phi_{n_p}}$ cells. This will, of course, change the manipulation graph connectivity and problems requiring the simultaneous manipulation of all the movable objects will be correctly reported to be not solvable\footnote{Note that, if the projection of a cell on a lower dimensional contact manifold is empty, then the lifting process does not take place and adjacency may only be  provided by transition through higher dimensional contact submanifolds or on a same leaf of a contact submanifold.}.

Finally, consider the adjacency by transit paths, i.e., paths along which the robot is not in contact with any object. The main idea is the same as before. It is simpler because we have to consider only the foliation induced by transit paths. The leaves of the foliation are 2-dimensional. We consider the cell decomposition of each $GRASP_{\Phi_{n_p}}$ after the previously described cell refinement. We add an edge between two cells $c_1$ and $c_2$ belonging respectively to $GRASP_{\Phi_{n_i}}$ and $GRASP_{\Phi_{n_j}}$ if and only if the projection of either one of the two cells onto the other along the foliation by transit paths is not empty\footnote{Note that we are implicitly assuming that the robot dynamics is symmetric.}.

\noindent
We have then the following\\
{\em Theorem:}  There exists a manipulation path between two configurations in the free space if and only if these configurations retract on two cells belonging to the same connected component of the manipulation graph.

The proof follows the same principle as the proof in~\cite{DaLaAl:92} and~\cite{SiLaCoSa:04}.

Wrapping up, the manipulation graph nodes are cells of the $S_m$ strata refined through the projection and lifting process described before; adjacency in the contact space is provided by transfer paths between the refined cells; adjacency in the robot free space is provided by transit paths between the refined cells.

\begin{figure}[t]
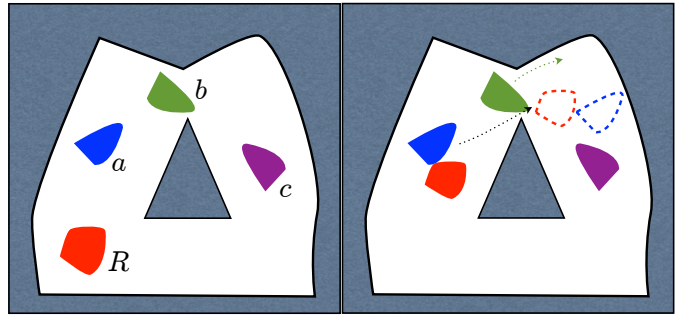

\begin{center}
\Ex
\caption{Example of a manipulation problem: the robot $R$ can move up to the three objects $a$, $b$, and $c$ (left) at the same time to achieve a manipulation task. The collision free path connecting two configurations of $GRASP_{a}$ and including the autonomous motion of $b$ (right) is an admissible path within $GRASP_{a}$ but not necessarily a manipulation path. Since $c$ is not moving, this path belongs to a leaf of $GRASP_{a}$.}
\label{fig:Ex}
\end{center}
\end{figure}

\begin{figure}[t]
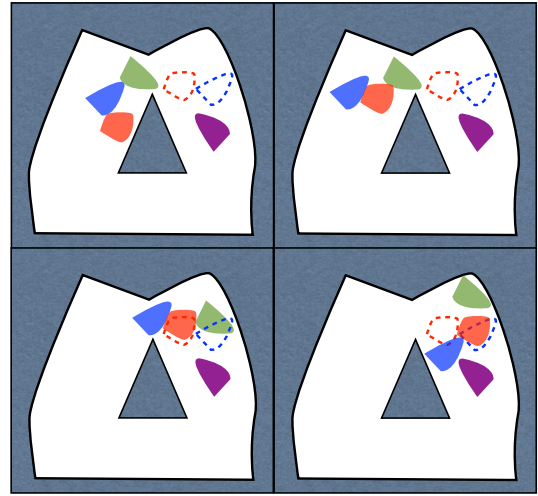

\begin{center}
\ExOne
\caption{To connect an initial and a desired configuration of $GRASP_{a}$ it is first necessary to establish a contact with $b$ (left), take $b$ to a convenient collision free configuration, i.e., move within $GRASP_{ab}$, and then reach the goal. Along this paths motion in contact is allowed.}
\label{fig:ExOne}
\end{center}
\end{figure}

\subsection{Example: $n=m=3$}
Consider the case of $n=3$ movable objects and a robot that can move up to $m=3$ objects together to achieve the planning task. The scenario is scketched in Fig.~\ref{fig:Ex} (left). 

In this case we have the following classes: $GRASP_{i}$, $i=\{a,b,c\}$, $GRASP_{ab}$, $GRASP_{ac}$, $GRASP_{bc}$, $GRASP_{abc}$, where the index denotes the object in contact with the robot.

The retraction of the cylindrical decomposition of the 8-dimensional configuration space on its boundary induces a cell decomposition of the (8-p)-dimensional ($p=1, \dots, 3$) contact submanifolds and this decomposition characterizes the connected components of each of the above listed classes $GRASP_{\Phi_{n_p}}$, $\Phi_{n_p} = \{a,b, c, ab, ac, bc, abc \}$. To refine the cell decomposition of each $GRASP_{\Phi_{n_p}}$ proceed as follows: 
\begin{enumerate}
\item merge the projections of $GRASP_{a}$ cell decomposition along the direction of its  foliations (corresponding to constant positions of the objects $b$ and $c$) onto $GRASP_{ab}$; this gives rise to a decomposition of this last class into many cells; these cells  can be reached by collision free paths in contact with $a$ that belong to the leaves of  $GRASP_{a}$; it is not guaranteed however that all the configurations in these cells can be reached by collision free paths in contact with $b$; this is guaranteed by a cell refinement illustrated in the next step;
\item repeat the analogous projection for $GRASP_{b}$ onto $GRASP_{ab}$; this further refines the decomposition of this last class after the projection of the first step; the cells in $GRASP_{ab}$ that appear to be at the basis of two cylinders containing cells of $GRASP_{a}$ and $GRASP_{b}$ can be reached by collision free paths respectively in contact with $a$ and $b$; any path within these cells belonging to a same leaf (i.e., the object $c$ remains in a same position) can be approximated by a collision free manipulation path; Fig.~\ref{fig:ExOne} illustrates through an example the adjacency by transfer path in class $GRASP_{a}$; to guarantee that any path within a cell in this class is collision free it is however necessary to refine the decomposition as illustrated below;
\item execute the steps analogous to 1. and 2. to refine the decomposition of $GRASP_{ac}$ and $GRASP_{bc}$;
\item merge in sequence the projections of $GRASP_{ab}$, $GRASP_{ac}$ and $GRASP_{bc}$ along their respective foliations onto $GRASP_{abc}$; this generates a decomposition of this last class into many elementary cells;
\item lift each elementary cell in $GRASP_{abc}$ to $GRASP_{ab}$, $GRASP_{ac}$ and $GRASP_{bc}$ along their respective foliations; each elementary cell in $GRASP_{abc}$ is at the basis of two cylinders containing cells of either $GRASP_{ab}$ and $GRASP_{ac}$ respectively or $GRASP_{ab}$ and $GRASP_{bc}$ or $GRASP_{ac}$  and $GRASP_{bc}$; these cells are nodes of the manipulation graph; 
\item lift each elementary cell in $GRASP_{ab}$, $GRASP_{ac}$ and $GRASP_{bc}$, obtained through the refinement in the previous step, to $GRASP_{a}$, $GRASP_{b}$ and $GRASP_{c}$ along their respective foliations; the cells in $GRASP_{a}$, $GRASP_{b}$ and $GRASP_{c}$ resulting from this refinement and belonging to at least one cylinder are nodes of the manipulation graph; the cell refinement is then completed.
\end{enumerate}

\section{Generality of the approach}
\label{sect:genapp}

To avoid formal complications, we have illustrated the decision process by making reference to a specific manipulation system.  The approach, however, is general enough to be applied to any dynamics of the manipulation system, any shape and any number of robots and objects. To support this generality claim, consider first the following synthetic description of the decision procedure.

\begin{enumerate}
\item Verify stratified controllability of the manipulation system.
\item If the system is stratified controllable, build the manipulation graph as described in 
Sect.~\ref{Sect:ManGraph}:
\begin{enumerate}
\item based on a cylindrical decomposition of ${\cal C}_{\rm free}$, determine the connected components of each $GRASP_{\Phi_{n_p}}$, $p = \{1,\dots, m\}$, $m=n$;
\item refine the cell decomposition of each $GRASP_{\Phi_{n_p}}$ through appropriate projections and lifting operations;
\item connect cells within each class $GRASP_{\Phi_{n_p}}$ and between classes $GRASP_{\Phi_{n_i}}$ and $GRASP_{\Phi_{n_j}}$  which are adjacent by transfer paths;
\item connect cells which are adjacent by transit paths;
\item if $m < n$, remove the nodes corresponding to classes $GRASP_{\Phi_{n_p}}$, $p > m$, together with all arcs corresponding to adjacency relations between cells that are based on the adjacency of $GRASP_{\Phi_{n_p}}$ cells.
\end{enumerate}
\item Search the manipulation graph for a solution; return failure if it does not exists.
\end{enumerate}

Note first that the use of a cylindrical decomposition as proposed in~\cite{Sc:83} to determine the connected components of the composite free configuration space allows robot, obstacles, objects and environment of any shape, provided that they posses a semi-algebraic geometry.
The chosen decomposition also allows to decompose the free configurations for multiple robots and objects. On the other hand, the stratified controllability test in~\cite{GoBu:01} can be repeatedly applied to multiple nested sequences of strata. If the top stratum in each sequence is different (as would be the case of multiple robots), then the test determines controllability for the union of the top strata. 
In addition, the test can be used with robots of any kinematic architecture. In principle, also the nature of the contact could be included in testing the controllability. In this case, however, we need to define other kind of adjacencies in addition to adjacency by transfer and transit paths characterising the dynamics of a rigid grasp.
 Hence, we argue that it is possible to build the manipulation graph for obstacles of any shape, in 2D and 3D workspaces, with multiple robots, possibly multi-articulated.

In fact, at the core of the decidability procedure is the controllability of the system. If the manipulation system is not controllable in the sense specified in Sect.~\ref{Sect:Controllability}, we are not able to conclude about decidability.

\section{Conclusion}
\label{Sect:Conclusion}

\noindent
We have proposed in this paper a decision procedure for the problem of planning the motion of systems with stratified configuration space.  Manipulation, climbing, walking, legged or multi-contact locomotion planning fall in the class of motion planning problems that can be handled with the presented approach. 

The decision process relies on the cylindrical decomposition of the composite (i.e., robot and movable objects) free configuration space and the construction of a manipulation graph whose nodes are cells in the composite free configuration space. Cell refinement and adjacency is determined by the stratified controllability property of the system on leaves of the contact configuration space. Within each cell in these submanifolds, paths in contact can be ({\em reduced}) to manipulation paths remaining in the free configuration space. 

Although illustrated for a particular manipulation planning problem, the tools and approach adopted are general enough to include a much wider class of problems as discussed in Sect.~\ref{sect:genapp}.

\bibliographystyle{IEEEtran}
\bibliography{IEEEabrv,DecManip}

\end{document}